\documentclass{article}

\usepackage{PRIMEarxiv}
\usepackage[utf8]{inputenc}
\usepackage[T1]{fontenc}
\usepackage{amsmath,amssymb,amsfonts}
\usepackage{booktabs}
\usepackage{graphicx}
\usepackage{microtype}
\usepackage{multirow}
\usepackage{nicefrac}
\usepackage{url}
\usepackage{xcolor}
\usepackage[numbers,sort&compress]{natbib}
\usepackage{hyperref}
\usepackage{authblk}
\usepackage{fancyhdr}
\usepackage{enumitem}
\usepackage{array}
\usepackage{caption}
\usepackage{float}

\graphicspath{{figures/}}
\hypersetup{
  colorlinks=true,
  citecolor=teal!55!black,
  linkcolor=teal!55!black,
  urlcolor=teal!55!black,
  pdftitle={HDND: Hierarchical Dynamic Neural Decoding for Multilingual Word and Character Retrieval from Non-Invasive Brain Recordings},
  pdfauthor={Yueyang Li, Shuran Chen, Wai Ting Siok, and Nizhuan Wang},
  pdfsubject={Multilingual word retrieval from non-invasive EEG/MEG},
  pdfkeywords={neural decoding, EEG, MEG, multilingual language decoding, word retrieval}
}
\setlist{nosep,leftmargin=1.5em}
\newcommand{\xneuro}{\mathbf{X}}

\newcommand{\zf}{\mathbf{z}^{f}}
\newcommand{\zp}{\mathbf{z}^{p}}
\newcommand{\es}{\mathbf{e}^{s}}
\newcommand{\ef}{\mathbf{e}^{f}}
\newcommand{\cosim}{\operatorname{cos}}
\newcommand{\loss}{\mathcal{L}}
\newcommand{\vocab}{\mathcal{V}}
\newcommand{\pp}{\ensuremath{\,\mathrm{pp}}}

\title{HDND: Hierarchical Dynamic Neural Decoding for Multilingual Word/Character Retrieval from Non-Invasive Brain Recordings}

\author[1]{Yueyang Li}
\author[1]{Shuran Chen}
\author[1,*]{Wai Ting Siok}
\author[1,*]{Nizhuan Wang}
\affil[1]{Department of Language Science and Technology, The Hong Kong Polytechnic University, Hung Hom, Kowloon, Hong Kong SAR, China}
\affil[*]{Corresponding authors: wai-ting.siok@polyu.edu.hk; wangnizhuan1120@gmail.com}

\begin{document}
\maketitle
\begin{abstract}
While deep learning has enabled language decoding from intracranial brain recordings, extending this capability to non-invasive recordings remains an open challenge. Decoding individual words from non-invasive brain recordings is particularly difficult, as word-level neural evidence is weak, temporally distributed, and entangled with acoustic, lexical, and semantic structure. Existing retrieval pipelines often collapse these factors into a single representation, potentially discarding information available at intermediate temporal scales. Here, we introduce Hierarchical Dynamic Neural Decoding (HDND), a hierarchical dynamic decoding framework that treats word decoding as structured refinement rather than flat label retrieval. HDND combines intermediate neural representations, contextual semantic predictions, and, for selected reading conditions, an auxiliary character-form objective. We evaluate HDND across nine conditions drawn from seven electroencephalography (EEG) and magnetoencephalography (MEG) datasets spanning English, Dutch, Mandarin, and Cantonese listening, reading, and reading-aloud conditions. Across the nine-condition word-retrieval benchmark, HDND yields a higher participant-averaged balanced Top-10 point estimate than the matched contextual word-decoding baseline in every condition and achieves the highest mean among all compared methods in eight of nine conditions. Across the same nine matched conditions, HDND also yields higher token-micro and pooled word-macro Top-10 point estimates in every setting. Sentence retrieval favors HDND in eight of nine conditions, while auditory speech-segment retrieval is mixed across the six listening conditions. These results show that hierarchical residual refinement can improve multilingual word retrieval from heterogeneous non-invasive brain recordings.
\end{abstract}


\section{Introduction}
\label{sec:introduction}

Language brain--computer interfaces (BCIs) have advanced rapidly with intracranial recordings, enabling speech and text neuroprostheses with practical rates and vocabularies for people with severe paralysis \cite{moses2021neuroprosthesis,willett2023speech,zhang2026imagined}. However, these systems require implanted electrodes and specialized clinical procedures, which limits their applicability to certain individuals and their viability for long-term use because of signal degradation and infection risk. Non-invasive electroencephalography (EEG) and magnetoencephalography (MEG) avoid surgery and capture neural dynamics at language-relevant timescales, but their signals have a lower signal-to-noise ratio, are spatially mixed (particularly for EEG), and are highly variable across participants and recording systems \cite{baillet2017magnetoencephalography,michel2019eeg,defossez2023decoding,li2025freqdgt}. Recent work has nevertheless advanced non-invasive decoding toward natural language. Semantic decoding using functional magnetic resonance imaging (fMRI) has been shown to reconstruct continuous language based on contextual meaning, though with limited temporal resolution \cite{tang2023semantic,zhang2026linguistics}. BrainMAGIC aligned EEG and MEG activity with speech representations for speech-segment retrieval \cite{defossez2023decoding}, whereas BrainAI extended neural-to-representation alignment to individual words using word-onset-aligned neural windows and fixed text embeddings \cite{dascoli2025words}. Together, these studies support representation alignment as a promising framework for non-invasive language decoding.

Individual-word retrieval nevertheless presents two coupled challenges. First, mapping an entire neural window to a single semantic embedding creates a representational bottleneck: neural responses can contain rapidly varying perceptual or subword cues together with more sustained lexical and contextual information \cite{brodbeck2018rapid,heilbron2022hierarchy}. Second, open-vocabulary retrieval benefits from an output space that shares statistical structure across words without reducing every distinction to semantic similarity. This requirement becomes more pronounced across writing systems. Alphabetic words are built from a small inventory of letters that map onto phonemes, so letter sequences recur across words. Chinese words are also combinatorial, but their units are characters, which map onto morphemes and syllables rather than phonemes. Characters may either stand alone as words or combine with others to form multisyllabic words. In contrast to English text, Chinese text lacks explicit word boundaries, so segmentation depends on language- and corpus-specific conventions \cite{sproat2003segmentation,lam2024cantonese}. Under these segmentation conventions, a lexical retrieval unit in Chinese (including Mandarin and Cantonese) may correspond to a single-character or multi-character word. Throughout this manuscript, the term \emph{word/character retrieval} refers to retrieval over segmentation-defined lexical units across writing systems rather than to two independent decoding tasks. Character n-grams, where \textit{character} denotes any written symbol (letters in alphabetic scripts; characters in Chinese), provide a common form-level representation for alphabetic strings and segmented Chinese lexical units, allowing lexical items to share surface components rather than acting as unrelated class identifiers \cite{bojanowski2017subword,vania2017characters,zhu2019subword,he2026toward}.

\begin{figure*}[t]
    \centering
    \includegraphics[width=\textwidth]{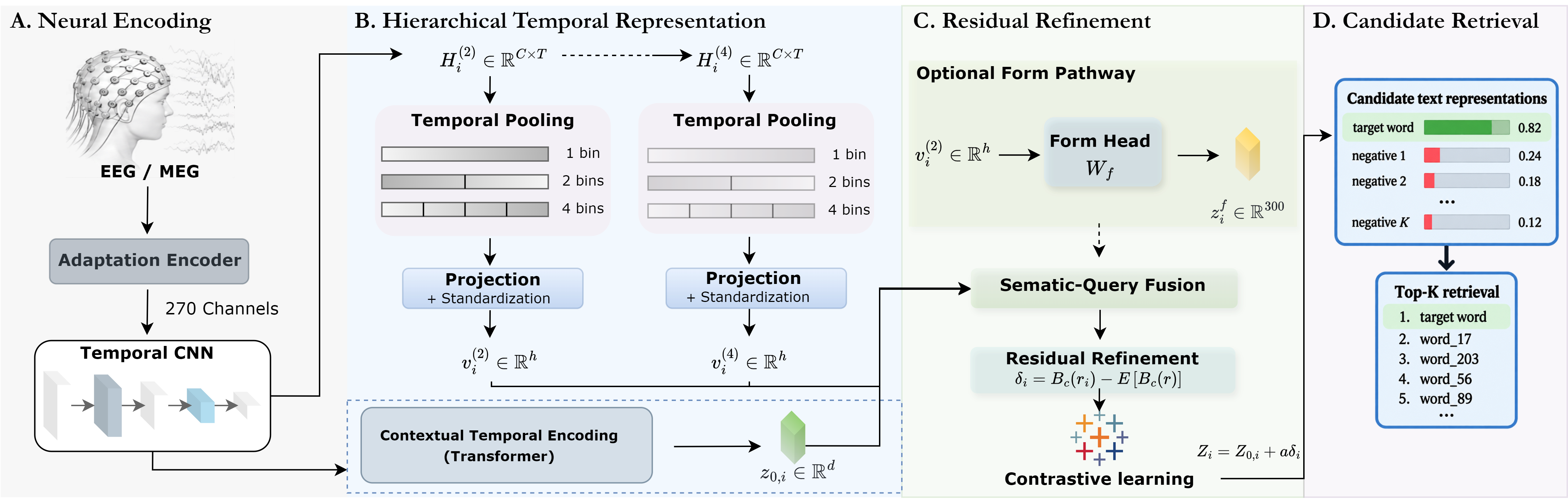}
    \caption{Architecture of the proposed HDND model. Neural recordings are mapped to a common sensor space and encoded by a temporal CNN. The outputs of the second and fourth GLU stages are pooled into one, two, and four temporal bins. A query derived from the contextual prediction weights the two depth-specific representations, and the fused evidence produces a bounded correction to the contextual prediction. The form pathway is considered only in reading conditions.}
    \label{fig:architecture}
\end{figure*}

We introduce \emph{Hierarchical Dynamic Neural Decoding} (HDND), a framework for word/character retrieval from heterogeneous non-invasive recordings (Fig.~\ref{fig:architecture}). HDND retains the geometry-aware and contextual backbone of BrainAI while exposing intermediate neural features at multiple temporal resolutions. A semantic query weights representations from two encoder depths, and their contribution is incorporated through a bounded residual. In reading conditions, an optional character-form objective provides additional supervision for surface structure. Across seven datasets and nine matched conditions, HDND produces a higher balanced Top-10 word-retrieval point estimate than BrainAI in every condition and the highest mean among the compared methods in eight.

\section{Related Work}
\label{sec:related}

Distributed language representations provide a shared space for relating linguistic content to neural activity. Early fMRI studies showed that semantic features could predict responses to individual concepts and generalize to unseen words \cite{mitchell2008predicting}, with subsequent work extending this paradigm to natural reading and continuous speech \cite{wehbe2014simultaneously,huth2016natural}. Corresponding semantic information has also been identified in temporally resolved EEG and MEG responses \cite{murphy2010detecting,broderick2018electrophysiological,toneva2020modeling}. More recently, pretrained speech and language models have supplied richer representations for modeling fMRI, intracranial recordings, EEG, and MEG \cite{caucheteux2021hierarchy,caucheteux2022deep,caucheteux2022brains,goldstein2024alignment,millet2022toward}. Neural language research has consequently shifted from encoding analysis toward direct decoding. Semantic content has been reconstructed from fMRI \cite{tang2023semantic}, and contrastive alignment with self-supervised speech representations has enabled speech-segment retrieval from EEG and MEG \cite{defossez2023decoding}. This approach was subsequently extended to individual-word decoding by predicting pretrained word representations from word-aligned neural responses \cite{dascoli2025words}. Compared with earlier linear approaches \cite{mitchell2008predicting,toneva2020modeling,caucheteux2022brains}, recent models benefit from multi-participant training and temporal contextualization \cite{defossez2023decoding,dascoli2025words}. Long-context pretraining and the decoding of typed language production have further broadened the scope of non-invasive language decoding \cite{jayalath2026megxl,levy2026noninvasive}.

Most word-level decoders, however, are optimized primarily against semantic embeddings. Language processing spans multiple scales, from acoustic and phonological structure to lexical form and contextual meaning \cite{brodbeck2018rapid,caucheteux2021hierarchy,millet2022toward,caucheteux2023predictive}. Consistent with this hierarchy, decoded representations retain information about word length, boundary letters, and part of speech, with some effects differing between reading and listening \cite{dascoli2025words}. Character- and subword-based representations offer a direct means of preserving form-level regularities across lexical items \cite{bojanowski2017subword}. Evidence from hierarchical decoding likewise suggests that acoustic, speech-model, and contextual language representations provide complementary rather than interchangeable supervision \cite{wang2026hierarchical}. A parallel challenge is maintaining useful representations across heterogeneous recordings. EEG and MEG vary substantially in sensor layout, acquisition system, participant anatomy, and experimental protocol. Convolutional architectures provide strong within-domain baselines \cite{lawhern2018eegnet,schirrmeister2017deep}, whereas spatial sensor mappings and subject-specific transformations improve multi-participant decoding \cite{defossez2023decoding,dascoli2025words}. Self-supervised methods such as BENDR, BIOT, and EEG2Rep further promote transfer across datasets and tasks \cite{kostas2021bendr,yang2023biot,foumani2024eeg2rep}. HDND builds on this work by integrating intermediate neural evidence, optional form supervision, and contextual semantic refinement within a common multi-resolution architecture.

\section{Hierarchical Dynamic Neural Decoding (HDND)}
\label{sec:method}

\subsection{Overview}

HDND predicts linguistic representations from word-aligned EEG and MEG recordings (Fig.~\ref{fig:architecture}). It augments a contextual neural decoder with two intermediate encoder representations, each summarized at three fixed temporal resolutions. These representations provide intermediate perceptual evidence for semantic refinement and, in reading conditions, for an optional character-form objective. Given neural example $i$, let $\xneuro_i\in\mathbb{R}^{C_i\times T}$ denote the recording aligned to target lexical unit $w_i$, where $C_i$ is the number of recorded neural channels and $T$ is the number of temporal samples in the input window. Let $\vocab$ denote the candidate retrieval vocabulary and let $\es(w)\in\mathbb{R}^{d}$ denote the fixed semantic embedding of candidate lexical unit $w\in\vocab$, where $d=1024$ is the semantic embedding dimensionality used in the present experiments. The model predicts a representation in this semantic space and compares it with the candidate embeddings. The sensor-alignment, temporal-encoding, and sentence-context modules follow the full word-decoding pipeline of \citet{dascoli2025words}. HDND retains this backbone and learns a residual refinement from intermediate and contextual neural features.

\subsection{Adaptation Encoder}

We adopt the sensor-position and subject-adaptation modules of \citet{dascoli2025words}. Fourier-encoded sensor coordinates and learned queries determine attention weights over the recorded channels. The resulting signals are mapped to $C_0=270$ latent channels, projected to 512 channels, and passed through a subject-specific linear transformation. Here, $C_0$ denotes the common number of latent channels after sensor adaptation. This front end places heterogeneous sensor layouts in a common representational space while accommodating participant variability within each cohort.

\subsection{Contextual Temporal Encoding}
The neural encoder comprises five temporal convolutional stages with dilated convolutions and residual connections. The first four stages use 160 output channels, whereas the fifth stage maps the representation to 1,024 dimensions. Gated linear units (GLUs) are applied after the second and fourth convolutional stages, whose outputs are retained for the hierarchical temporal pathway. Attention-based temporal aggregation maps each word window to a
1,024-dimensional vector. A 16-layer, 16-head Transformer then combines the word-level neural vectors within each annotated sentence and produces the normalized contextual prediction $\mathbf{z}_{0,i}$. Candidate words are embedded independently, and contextual information is introduced only by the neural Transformer. Representations are extracted from the middle layer of the pretrained text encoder, and subtoken representations are averaged when a candidate is divided into multiple tokens.

\subsection{Hierarchical Temporal Pyramid}
\label{sec:temporal-pyramid}

A single temporal summary may suppress information that remains available at intermediate encoder stages. We therefore retain the outputs of the second and fourth GLU stages, $\mathbf{H}^{(2)}_i$ and $\mathbf{H}^{(4)}_i$, which differ in convolutional depth and effective temporal context. For each depth, the temporal feature map is adaptively pooled at three fixed resolutions: one global bin, two coarse temporal bins, and four finer temporal bins:
\begin{equation}
 \mathbf{p}^{(\ell)}_i =
 \operatorname{Concat}_{b\in\{1,2,4\}}
 \operatorname{vec}\!\left(
 \operatorname{AvgPool}_{b}(\mathbf{H}^{(\ell)}_i)\right),
 \qquad \ell\in\{2,4\}.
 \label{eq:pyramid}
\end{equation}
Here, $i$ indexes neural examples, $\ell\in\{2,4\}$ indexes the selected GLU stage, and $\mathbf{H}^{(\ell)}_i$ denotes the corresponding intermediate temporal feature map. The variable $b\in\{1,2,4\}$ denotes the number of adaptive temporal bins. $\operatorname{AvgPool}_{b}(\cdot)$ performs adaptive average pooling into $b$ temporally ordered bins, $\operatorname{vec}(\cdot)$ vectorizes the resulting feature map, and $\operatorname{Concat}(\cdot)$ concatenates the vectors obtained at the three temporal resolutions. The resulting vector $\mathbf{p}^{(\ell)}_i$ is therefore the multi-resolution temporal-pyramid representation for encoder depth $\ell$.

The resulting representation preserves coarse temporal order while providing a fixed-dimensional interface across encoder depths. Because each selected GLU stage has 160 channels, concatenating the $1$-, $2$-, and $4$-bin summaries yields $160\times(1+2+4)=1,120$ features for each depth. Importantly, these temporal resolutions are fixed and concatenated; the dynamic weighting introduced below operates across encoder depths rather than across individual temporal bins.

\subsection{Perception, Form, and Semantic Pathways}
\label{sec:factors}

HDND assigns complementary roles to intermediate neural evidence, surface form, and semantics. The optional form pathway regularizes this representation in reading conditions, whereas the semantic pathway determines how intermediate evidence modifies the contextual prediction.

\paragraph{Intermediate neural representation.}
The pyramid vectors are standardized using training-set statistics and projected into a common space:
\begin{equation}
 \mathbf{v}^{(\ell)}_i =
 \tanh\!\left[
 \mathbf{P}_{\ell}
 \left(\frac{\mathbf{p}^{(\ell)}_i-\boldsymbol{\mu}_{\ell}}
 {\boldsymbol{\sigma}_{\ell}}\right)\right],
 \qquad \zp_i=\mathbf{v}^{(2)}_i.
 \label{eq:perceptual}
\end{equation}
Here, $\boldsymbol{\mu}_{\ell}\in\mathbb{R}^{1120}$ and $\boldsymbol{\sigma}_{\ell}\in\mathbb{R}^{1120}$ denote the feature-wise location and scale statistics estimated exclusively from the training partition for encoder depth $\ell$ and subsequently fixed for validation and test evaluation. Subtraction and division are performed element-wise. $\mathbf{P}_{\ell}\in\mathbb{R}^{h\times1120}$ is the learned projection matrix, where the projected width $h$ is 64 or 128. The projected representation $\mathbf{v}^{(\ell)}_i\in\mathbb{R}^{h}$ denotes the standardized intermediate neural representation at depth $\ell$, and $\zp_i=\mathbf{v}^{(2)}_i\in\mathbb{R}^{h}$ denotes the perceptual representation supplied to the auxiliary form head. It is learned through the retrieval objective and, for reading data when selected, the form objective; no frame-level phoneme labels or connectionist temporal classification (CTC) supervision are used.

\paragraph{Word/Character form in reading conditions.}
For each boundary-marked lexical string, the set of unique character $n$-grams of lengths 2--6 is constructed. Each $n$-gram is assigned a deterministic 300-dimensional Gaussian projection generated with a fixed random seed, and the projections associated with a lexical unit are summed and L2-normalized. The resulting prototype bank is centered using the mean form representation of the training labels and normalized again. We denote the resulting fixed prototype for target lexical unit $w_i$ by $\ef(w_i)\in\mathbb{R}^{300}$. A bias-free form head predicts this target from the perceptual representation:
\begin{equation}
 \zf_i=\mathbf{W}_f\zp_i,\qquad
 \loss_{\mathrm{form}}=
 \frac{1}{N}\sum_{i=1}^{N}
 \left[1-\cosim\!\left(\zf_i,\ef(w_i)\right)\right].
 \label{eq:form}
\end{equation}
The predicted form vector is mapped to the semantic space before integration. Character-form supervision is considered only for reading conditions, where its inclusion is determined on validation data. For listening conditions, the
form loss and form-to-semantic residual are disabled. Both settings are reported under the same HDND name because they instantiate the same task-adapted framework.

\paragraph{Semantic-query fusion.}
The contextual prediction supplies a query for weighting the two encoder-depth representations. Let $\mathbf{q}_i\in\mathbb{R}^{h}$ denote the semantic query obtained by feature-wise standardizing $\mathbf{z}_{0,i}$ using statistics estimated exclusively from the training partition, applying a learned bias-free linear projection from $\mathbb{R}^{d}$ to $\mathbb{R}^{h}$, and then applying a $\tanh$ nonlinearity. The depth weights and fused representation are
\begin{equation}
 a_{i\ell}=\operatorname{softmax}_{\ell}
 \left(\frac{\mathbf{q}_i^\top\mathbf{v}^{(\ell)}_i}{\sqrt{h}}\right),
 \qquad
 \mathbf{u}_i=
 \sum_{\ell\in\{2,4\}}a_{i\ell}\mathbf{v}^{(\ell)}_i
 +\eta\mathbf{q}_i,
 \label{eq:gating}
\end{equation}
where $a_{i\ell}$ is the normalized sample-specific weight assigned to encoder depth $\ell$, such that $\sum_{\ell\in\{2,4\}}a_{i\ell}=1$. $\mathbf{u}_i\in\mathbb{R}^{h}$ denotes the fused intermediate representation, and $\eta\in\{0,1\}$ controls the direct query connection.

The semantic head produces a correction, augmented by the form projection when the form pathway is enabled:
\begin{equation}
 \mathbf{r}_i=\mathbf{W}_s\mathbf{u}_i
 +\mathbb{I}_{\mathrm{form}}\mathbf{W}_{fs}\zf_i,
 \label{eq:factor-residual}
\end{equation}
where $\mathbf{W}_s\in\mathbb{R}^{d\times h}$ maps the fused intermediate representation into the semantic space, $\mathbf{W}_{fs}\in\mathbb{R}^{d\times300}$ maps the predicted form representation into the same space, and $\mathbf{r}_i\in\mathbb{R}^{d}$ is the resulting pre-bounding residual correction. The indicator $\mathbb{I}_{\mathrm{form}}\in\{0,1\}$ equals one when the form pathway is enabled and zero otherwise. The retrieval decision therefore remains in the same fixed text-embedding space while depending on sample-specific intermediate neural evidence and, when appropriate, character-form information.

\subsection{Residual Refinement and Retrieval}

To preserve a stable reference to the contextual prediction, the pre-centering correction is smoothly bounded and its training-set mean is removed:
\begin{equation}
 \boldsymbol{\delta}_i=
 \mathcal{B}_c(\mathbf{r}_i)-
 \mathbb{E}_{j\in\mathrm{train}}
 \!\left[\mathcal{B}_c(\mathbf{r}_j)\right],
 \qquad
 \mathcal{B}_c(\mathbf{r})=
 \frac{c\mathbf{r}}{\sqrt{c^2+\lVert\mathbf{r}\rVert_2^2}}.
 \label{eq:bounded-residual}
\end{equation}
Here, $\mathcal{B}_c(\cdot)$ denotes the smooth residual-bounding operator, $c>0$ is its norm-control parameter, and $\lVert\cdot\rVert_2$ denotes the Euclidean norm. The index $j$ runs over training examples, and $\mathbb{E}_{j\in\mathrm{train}}[\cdot]$ denotes the empirical mean computed from training examples only. The bound $c$ is 0.15 or 0.25. The empirical mean is estimated only from training examples and is fixed for validation and test evaluation. Because $\lVert\mathcal{B}_c(\mathbf{r})\rVert_2<c$, the centered residual $\boldsymbol{\delta}_i\in\mathbb{R}^{d}$ has norm below $2c$. The output matrices $\mathbf{W}_s$ and $\mathbf{W}_{fs}$ are initialized to zero, so optimization starts from the contextual prediction.

The final prediction and candidate score are
\begin{equation}
 \mathbf{z}_i=\mathbf{z}_{0,i}+\alpha\boldsymbol{\delta}_i,
 \qquad
 S_i(w)=\mathbf{z}_i^\top\es(w),
 \label{eq:retrieval}
\end{equation}
where $\alpha$ is selected on the validation set and $\es(w)$ is the L2-normalized semantic embedding of candidate lexical unit $w$. Candidates are ranked by $S_i(w)$. The refined prediction $\mathbf{z}_i$ is not explicitly renormalized after residual addition. However, because its norm is constant across candidates for a given example,
dot-product scoring induces the same candidate ordering as cosine similarity. The form pathway affects ranking only through the joint residual and is not used as a separate candidate score.

\subsection{Training}
\label{sec:training}

Training proceeds in two stages. The geometry-aware sensor encoder, temporal CNN, temporal aggregation module, and sentence Transformer are first trained jointly from random initialization to align contextual neural predictions with target word embeddings. These components are then frozen, and the temporal-pyramid projections, semantic-query module, and residual output heads are optimized.

The refinement objective is
\begin{equation}
\begin{aligned}
 \loss &=
 \loss_{\mathrm{ret}}
 +\lambda_{\mathrm{KL}}\loss_{\mathrm{anchor}}
 +\lambda_f\loss_{\mathrm{form}}
 +\lambda_p\loss_{\mathrm{pair}},\\
 \loss_{\mathrm{pair}}
 &=
 \frac{1}{B}\sum_{i=1}^{B}
 \operatorname{softplus}\!\left[
 -\tau
 \left(\boldsymbol{\delta}_i-
 \tilde{\boldsymbol{\delta}}_i\right)^\top
 \es(w_i)
 \right].
\end{aligned}
\label{eq:loss}
\end{equation}
Here, $\loss_{\mathrm{ret}}$ is candidate cross-entropy, optionally weighted by training-word frequency. The anchor term constrains the refined candidate distribution to remain close to the contextual prediction through $D_{\mathrm{KL}}(p_{0,i}\Vert p_i)$. The form loss is defined in Eq.~\ref{eq:form} and is enabled only for selected reading-condition
configurations. In the paired-feature term, $B$ denotes the minibatch size, $\tilde{\boldsymbol{\delta}}_i$ denotes the residual obtained after shuffling the intermediate hierarchical features while retaining the contextual prediction for example $i$, and $\tau$ denotes the fixed logit scale used during refinement optimization. The loss therefore encourages the correctly aligned residual to assign the target lexical unit a higher score than its shuffled counterpart. For listening conditions, $\lambda_f=0$ and the form-to-semantic projection is omitted.

The refinement module is trained with AdamW using a batch size of 1,024, weight decay of 0.01, cosine learning-rate decay, and gradient-norm clipping at 1.0. We consider projected widths of 64 and 128, residual bounds of 0.15 and 0.25, and integration strengths of 0.125, 0.25, 0.5, and 1.0. Model configuration, checkpoint, and integration strength are selected exclusively on validation data and subsequently applied unchanged to the test set. The complete configuration and validation search space are summarized in Table~\ref{tab:hyperparameters}.

\section{Experimental Protocol}
\label{sec:experiments}

\subsection{Datasets}

We evaluate seven EEG and MEG datasets spanning English, Dutch, Mandarin, and Cantonese (Table~\ref{tab:conditions}). Together, these datasets define nine primary experimental conditions covering auditory listening, silent reading, and reading-aloud paradigms, with stimuli ranging from continuous narratives to controlled sentence presentation. A separate model is trained and evaluated for each condition. Armeni, Broderick, and Gwilliams provide English recordings acquired during naturalistic speech comprehension \cite{armeni2022dataset,broderick2018electrophysiological,gwilliams2023megmasc}. Armeni contains extended within-participant MEG recordings, Gwilliams includes story-listening MEG from a larger cohort, and Broderick contributes audiobook-listening EEG. Schoffelen provides Dutch MEG from separate listening and reading groups, denoted Schoff-L and Schoff-R \cite{schoffelen2019mous}. ChineseEEG2.0 contains Mandarin EEG acquired during passive listening and reading aloud of the same narrative materials \cite{chen2026chineseeeg2}; the two tasks are treated as distinct experimental conditions. LPPHK provides Cantonese EEG from participants listening to \emph{Le Petit Prince} \cite{momenian2024lpphk}. Nieuwland contains English sentences presented word by word using a rapid serial visual presentation (RSVP) paradigm \cite{nieuwland2018replication}.

\begin{table*}[t]
\centering
\caption{Datasets and experimental conditions in the primary benchmark. $N$ denotes the number of participants retained in the reported analysis. L and R are retained as dataset-specific labels for the auditory and visually presented conditions, respectively. ``Yes'' in the final column indicates that character-form supervision was included in the validation search space.}
\label{tab:conditions}
\scriptsize
\resizebox{\textwidth}{!}{%
\begin{tabular}{llllcccl}
\toprule
Dataset & Language & Task & Stimuli & Device & $N$ & Sensors & Used Form \\
\midrule
Armeni \cite{armeni2022dataset} & English & Listen & Narratives & MEG & 3 & 298 & No \\
Broderick \cite{broderick2018electrophysiological} & English & Listen & Audiobook & EEG & 19 & 128 & No \\
Gwilliams \cite{gwilliams2023megmasc} & English & Listen & Stories & MEG & 27 & 208 & No \\
Schoff-L \cite{schoffelen2019mous} & Dutch & Listen & Sentences & MEG & 96 & 273 & No \\
Schoff-R \cite{schoffelen2019mous} & Dutch & Read & Sentences & MEG & 99 & 273 & Yes \\
ChineseEEG2.0-L \cite{chen2026chineseeeg2} & Mandarin & Passive listening & Narratives & EEG & 7 & 128 & No \\
ChineseEEG2.0-R \cite{chen2026chineseeeg2} & Mandarin & Read aloud & Narratives & EEG & 4 & 128 & Yes \\
LPPHK \cite{momenian2024lpphk} & Cantonese & Listen & Audiobook & EEG & 49 & 64 & No \\
Nieuwland \cite{nieuwland2018replication} & English & Read (RSVP) & Sentences & EEG & 295 & 63 & Yes \\
\bottomrule
\end{tabular}%
}
\end{table*}

\subsection{Preprocessing and Data Partitions}

Recordings are band-pass filtered between 0.1 and 40~Hz and subsequently resampled to 50~Hz with anti-alias filtering, yielding an effective upper frequency limit of 25~Hz. Robust-scaling statistics are estimated separately for each channel using the training partition only and are subsequently applied unchanged to the validation and test partitions. The scaled signals are then clamped to \([-5,5]\). Each example consists of a 3-s neural window beginning at the annotated word onset, and baseline correction is applied using the first 0.5~s of the extracted window. Sensor coordinates, participant identifiers, word-onset annotations, and sentence groupings are retained for model construction and evaluation. For the naturalistic conditions, train, validation, and test partitions follow an 80/10/10 split at the sentence level. Sentences are assigned by deterministic hashing so that repeated presentations of the same sentence across participants remain in the same partition. Participants may therefore contribute recordings to all three partitions, and the reported evaluation measures generalization to held-out linguistic content rather than zero-shot generalization to unseen participants. BrainAI and HDND use identical preprocessed recordings, data partitions, target representations, candidate sets, and evaluation code in every matched comparison.

\subsection{Models and Controls}

We compare six decoding approaches: Linear, Conv, EEGNet \cite{lawhern2018eegnet}, BrainMAGIC, BrainAI, and HDND. Linear and Conv denote the corresponding linear and convolutional baselines implemented under the matched evaluation protocol. For compactness, we use \emph{BrainMAGIC} to refer to the contrastive brain--speech decoding approach based on the brain module of \citet{defossez2023decoding}, and \emph{BrainAI} to refer to the convolutional neural network (CNN)-plus-sentence-Transformer contextual word-decoding pipeline of \citet{dascoli2025words}. These names are manuscript-specific shorthand used consistently across figures and tables. All within-condition comparisons use identical neural inputs, data partitions, target representations, candidate sets, and evaluation procedures. For each condition, BrainAI and HDND use the same trained contextual backbone, with HDND fitted as a refinement of the frozen backbone. Consistent with the matched BrainAI reporting convention, condition-level uncertainty is summarized across participants rather than across random initializations.

Two controls probe complementary sources of the refinement gain produced by HDND. The \emph{base-output-only} control removes the hierarchical temporal pathway and learns a residual correction using only the final contextual prediction. The \emph{shuffled-pyramid} control retains the hierarchical pathway but permutes the intermediate pyramid features within participant while preserving the correctly paired contextual prediction. The base-output-only control therefore tests how much improvement can be obtained through contextual remapping alone, whereas the shuffled-pyramid control tests whether the intermediate pathway depends on sample-specific neural information. Where reported, a frequency-only reference ranks the ten most frequent training words highest and does not use neural input.

\subsection{Retrieval Tasks and Metrics}

\paragraph{Word/character retrieval.} The primary lexical retrieval task operates over segmentation-defined candidate units. For alphabetic languages, these units correspond to words. For Mandarin and Cantonese, the language-specific segmentation procedure may yield either a single-character lexical item or a multi-character word. For notational simplicity, $w$ denotes a candidate lexical unit in all languages. For each neural prediction, candidate lexical units are ranked by cosine similarity in the fixed text-embedding space. Following the evaluation protocol of \citet{dascoli2025words}, cross-dataset comparisons use a fixed vocabulary containing the 250 most frequent eligible lexical units for each condition. Lexical units outside this reduced vocabulary are excluded from the primary Top-250 evaluation. Because the decoder predicts continuous pretrained lexical representations rather than a vocabulary-sized class distribution, lexical retrieval is not intrinsically restricted to labels observed during neural-network training, and lexical units absent from the training partition can be examined separately in zero-shot analyses.

The primary metric is single-trial participant-averaged balanced Top-10 accuracy. Let $s$ index participants, $i$ index evaluated lexical-unit occurrences, and $w$ index target lexical-unit types. Let $c_{s,i}\in\{0,1\}$ indicate whether the correct lexical unit for trial $i$ from participant $s$ appears among the ten highest-scoring candidates, let $\mathcal{I}_{s,w}$ denote the set of evaluated occurrences of target lexical unit $w$ for participant $s$, and let $\mathcal{W}^{\mathrm{eval}}_s$ denote the set of evaluated target lexical-unit types observed for that participant. We compute
\begin{equation}
A_{\mathrm{bal}}^{(s)}=
\frac{1}{|\mathcal{W}^{\mathrm{eval}}_s|}
\sum_{w\in\mathcal{W}^{\mathrm{eval}}_s}
\frac{1}{|\mathcal{I}_{s,w}|}
\sum_{i\in\mathcal{I}_{s,w}}c_{s,i},
\qquad
\bar A_{\mathrm{bal}}=
\frac{1}{|\mathcal{S}|}
\sum_{s\in\mathcal{S}}A_{\mathrm{bal}}^{(s)}.
\label{eq:balanced-top10}
\end{equation}
Here, $\mathcal{S}$ denotes the set of evaluated participants, $A_{\mathrm{bal}}^{(s)}$ is the balanced Top-10 accuracy for participant $s$, and $\bar A_{\mathrm{bal}}$ is the participant-averaged condition-level balanced Top-10 accuracy. Thus, trial outcomes are first averaged within lexical-unit type, lexical-unit accuracies are averaged within participant, and participant-level accuracies are then averaged within condition. Error bars report the standard error of the mean (SEM) across participants. Uniform random Top-10 retrieval among 250 candidates has an expected accuracy of 4\%.

For the auxiliary analyses reported in Appendix~\ref{app:full-results}, we additionally report token-micro and pooled word-macro Top-10 accuracy:
\begin{equation}
A_{\mathrm{micro}}=
\frac{1}{N_{\mathrm{eval}}}
\sum_{i=1}^{N_{\mathrm{eval}}}c_i,
\qquad
A_{\mathrm{word\text{-}macro}}=
\frac{1}{|\mathcal{W}_{\mathrm{obs}}|}
\sum_{w\in\mathcal{W}_{\mathrm{obs}}}
\frac{1}{|\mathcal{I}_w|}
\sum_{i\in\mathcal{I}_w}c_i.
\label{eq:macro}
\end{equation}
Here, $N_{\mathrm{eval}}$ denotes the total number of evaluated lexical-unit occurrences pooled over the corresponding evaluation set, and $c_i\in\{0,1\}$ indicates whether the correct target for occurrence $i$ appears among the Top-10 predictions. $\mathcal{W}_{\mathrm{obs}}$ denotes the set of distinct target lexical-unit types observed among those evaluated occurrences, and $\mathcal{I}_w=\{i:w_i=w\}$ denotes the set of evaluated occurrence indices whose target lexical unit is $w$. Token-micro accuracy weights every evaluated occurrence equally, whereas pooled word-macro accuracy first averages occurrences within each observed lexical-unit type and then weights lexical-unit types equally. These pooled metrics are therefore complementary to, rather than interchangeable with, the participant-averaged balanced metric used for the primary nine-condition benchmark. Full-vocabulary auxiliary evaluation retains all eligible candidate types and test targets defined by the corresponding condition-level protocol; training-set occurrence counts can additionally be used to identify lexical units that are out of vocabulary with respect to neural training.

\paragraph{Sentence retrieval.} Sentence retrieval uses an endpoint-specific target representation and candidate set but follows the same general retrieval principle. A neural representation associated with each sentence is compared with fixed embeddings of candidate sentences using cosine similarity, and candidates are ranked according to the resulting scores. We report participant-averaged Top-10 sentence-retrieval accuracy under the matched sentence protocol. Because sentence retrieval uses a different target representation and candidate space from word retrieval, its absolute accuracy should not be interpreted on the same numerical scale as word-level Top-10 accuracy.

\paragraph{Auditory speech-segment retrieval.} Auditory speech-segment retrieval follows the segment-identification formulation of \citet{defossez2023decoding} and is evaluated only for conditions in which participants receive externally presented auditory speech. Each 3-s neural segment is paired with the temporally corresponding 3-s speech segment, whose target representation is obtained from the pretrained speech encoder used by the matched retrieval protocol. The decoder ranks the matched speech representation among held-out candidate segments, and Top-10 accuracy measures whether the correct segment appears among the ten highest-scoring candidates. This endpoint is therefore evaluated for Armeni, Broderick, Gwilliams, Schoff-L, ChineseEEG2.0-L, and LPPHK, while Schoff-R, ChineseEEG2.0-R, and Nieuwland are excluded because they do not contain an externally presented auditory-speech target. As with sentence retrieval, speech-segment accuracy is endpoint specific and is not numerically comparable to the word-retrieval metric.

\section{Results}
\label{sec:results}

\subsection{Word/Character Retrieval}

\begin{figure*}[t]
\centering
\includegraphics[width=\textwidth]{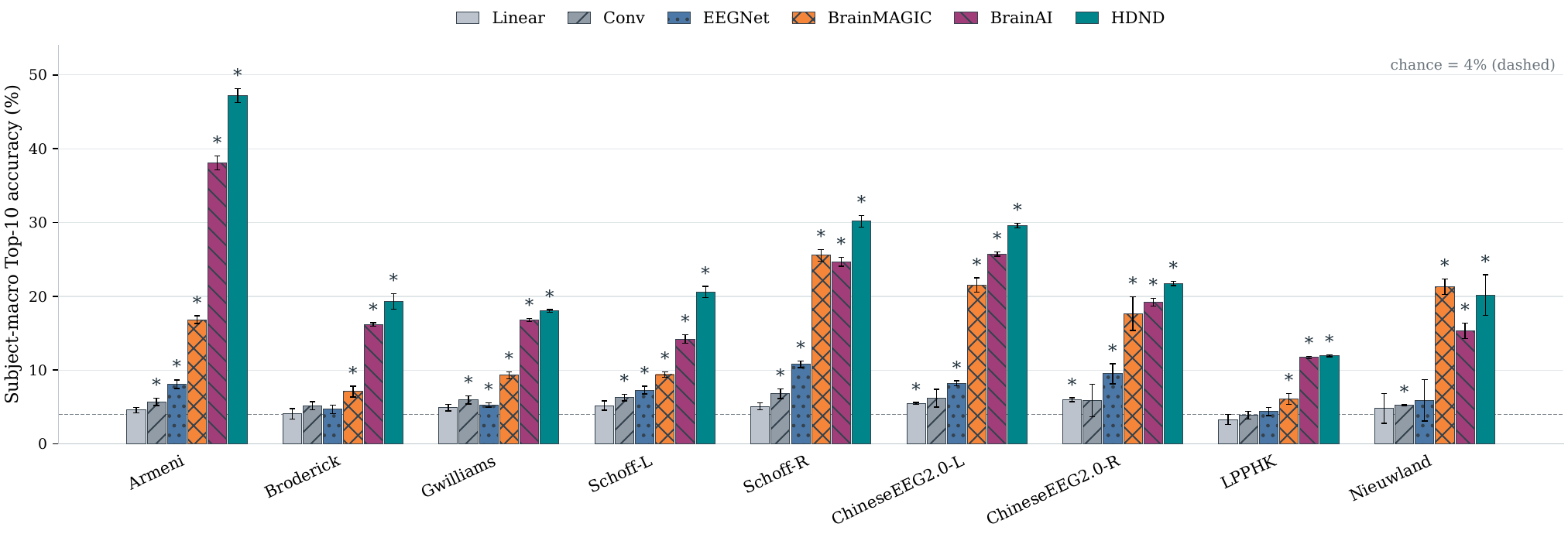}
\caption{Word/character retrieval across the nine primary conditions. Bars show participant-averaged single-trial balanced Top-10 accuracy over 250 candidates; error bars show SEM across participants. The dashed line denotes uniform-random retrieval (4\%). Stars indicate above-chance decoding across participants ($p<0.005$, one-sided test) and do not test differences between models.}
\label{fig:word-condition-bars}
\label{fig:primary-results}
\end{figure*}

We first evaluate word/character retrieval under the nine-condition protocol matched to BrainAI (Fig.~\ref{fig:word-condition-bars}). HDND yields a higher participant-averaged balanced Top-10 point estimate than BrainAI in every condition, with a mean difference of 4.09\pp. The largest absolute accuracy and the largest difference are observed for Armeni, where HDND reaches 47.19\% and exceeds BrainAI by 9.12\pp. Differences are also observed for both Schoffelen conditions (6.36\pp\ for Schoff-L and 5.52\pp\ for Schoff-R), indicating that the favorable direction is not confined to either listening or reading. Complete condition-level results are reported in Table~\ref{tab:full-word250}.

The additional Top-250 analyses report token-micro and pooled word-macro accuracy for the same nine primary conditions (Table~\ref{tab:main-results}). Token-micro accuracy weights every evaluated occurrence equally, whereas pooled word-macro accuracy gives equal weight to each observed lexical-unit type. The reported results show positive HDND--BrainAI differences for both metrics, although the magnitude of the difference varies across conditions.

\begin{table}[H]
\centering
\caption{Additional Top-250 word/character-retrieval metrics across the nine primary conditions. Values are token-micro and pooled word-macro percentages from the validation-selected model for each condition on the shared trained backbone. Differences are reported in percentage points (pp) and are computed before rounding.}
\label{tab:main-results}
\small
\begin{tabular}{lrrrrrr}
\toprule
& \multicolumn{3}{c}{Token-micro}
& \multicolumn{3}{c}{Word-macro} \\
\cmidrule(lr){2-4}\cmidrule(lr){5-7}
Condition & BrainAI & HDND & $\Delta$
& BrainAI & HDND & $\Delta$ \\
\midrule
Armeni
& 32.10 & 40.85 & +8.75
& 18.50 & 21.40 & +2.90 \\
Broderick
& 8.22 & 11.26 & +3.04
& 6.18 & 7.17 & +0.99 \\
Gwilliams
& 25.98 & 31.13 & +5.15
& 16.87 & 18.09 & +1.22 \\
Schoff-L
& 11.80 & 17.60 & +5.80
& 8.20 & 9.45 & +1.25 \\
Schoff-R
& 18.90 & 24.20 & +5.30
& 13.60 & 14.80 & +1.20 \\
ChineseEEG2.0-L
& 38.66 & 44.82 & +6.16
& 28.79 & 29.55 & +0.76 \\
ChineseEEG2.0-R
& 36.41 & 40.31 & +3.90
& 21.20 & 21.65 & +0.45 \\
LPPHK
& 13.48 & 15.64 & +2.16
& 11.72 & 11.95 & +0.23 \\
Nieuwland
& 15.52 & 20.86 & +5.33
& 11.12 & 13.11 & +1.99 \\
\bottomrule
\end{tabular}
\end{table}

\subsection{Core Component Analysis}

We next evaluate the core components of HDND under the same nine conditions used in the primary word/character-retrieval benchmark (Table~\ref{tab:core-ablations}). The base-output-only control removes the hierarchical temporal pathway and learns the residual correction from the contextual output alone, whereas the shuffled-pyramid control retains the hierarchical pathway but disrupts the correspondence between each trial and its intermediate features.

For Armeni, the shuffled-pyramid and base-output-only controls achieve balanced Top-10 accuracies of 43.89\% and 44.36\%, respectively, compared with 47.19\% for full HDND. For Schoff-L, the corresponding accuracies are 18.52\% and 19.76\%, compared with 20.58\% for HDND; for Schoff-R, they are 27.77\% and 28.53\%, compared with 30.19\% for HDND.

Full-vocabulary retrieval provides a complementary test of candidate-set dependence (Table~\ref{tab:full-vocabulary}). The analysis uses the same primary condition definitions while expanding the candidate bank beyond the fixed Top-250 vocabulary. Across the reported full-vocabulary results, HDND retains higher token-micro and pooled word-macro point estimates than BrainAI.

\subsection{Sentence Retrieval}

\begin{figure*}[t]
\centering
\includegraphics[width=\textwidth]{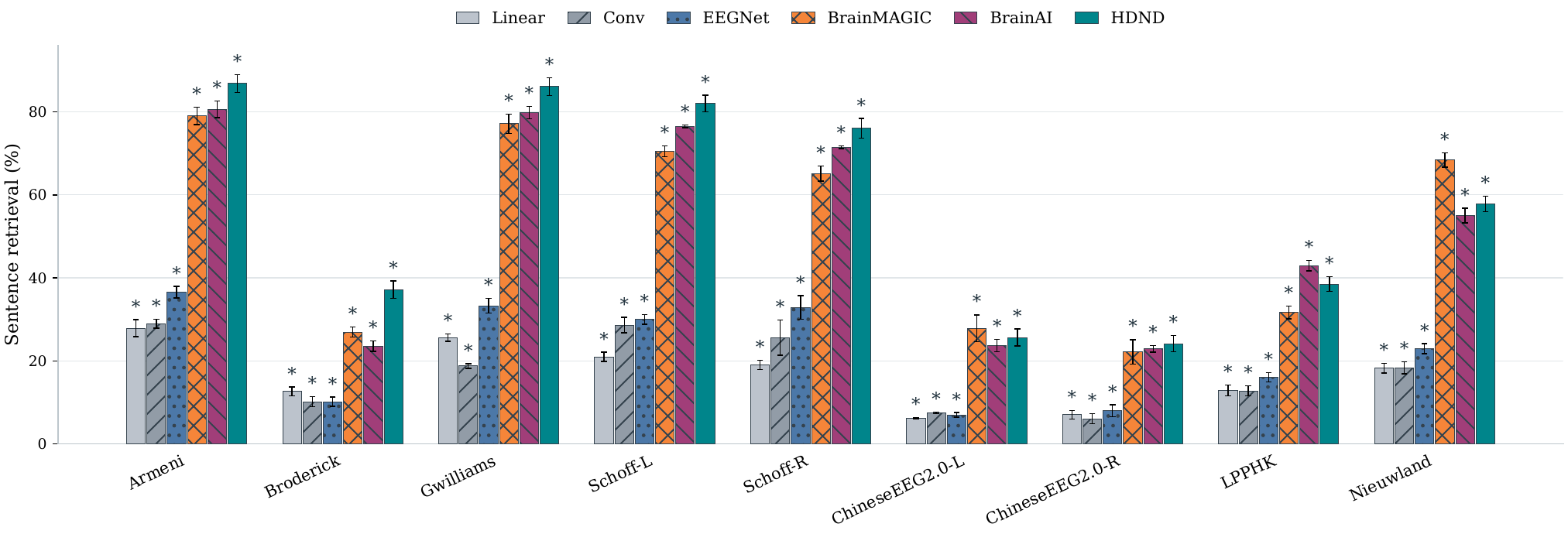}
\caption{Sentence retrieval across the nine primary conditions. Bars show participant-averaged Top-10 accuracy, and error bars show SEM across participants. Stars indicate above-chance decoding ($p<0.005$, one-sided test) and do not test between-model differences. Complete values are reported in Table~\ref{tab:full-sentence}.}
\label{fig:sentence-condition-bars}
\end{figure*}

Sentence retrieval asks whether the favorable lexical comparison extends to complete sentences (Fig.~\ref{fig:sentence-condition-bars}; complete values are reported in Table~\ref{tab:full-sentence}). HDND yields a higher point estimate than BrainAI in eight of the nine primary conditions, with a mean difference of 4.19\pp. The largest difference occurs on Broderick (13.60\pp), whereas LPPHK is the only condition favoring BrainAI ($-4.44\pp$). The relative difficulty of the Schoffelen conditions also changes with the retrieval unit: Schoff-R is easier for individual words, while Schoff-L is easier for sentences in both models.

The broader model ranking also changes across endpoints. BrainMAGIC leads on ChineseEEG2.0-L and Nieuwland, and BrainAI remains higher than HDND on LPPHK. Improved discrimination among lexical candidates therefore does not guarantee a stronger sentence-level representation.

\subsection{Auditory Speech-Segment Retrieval}

\begin{figure*}[t]
\centering
\includegraphics[width=\textwidth]{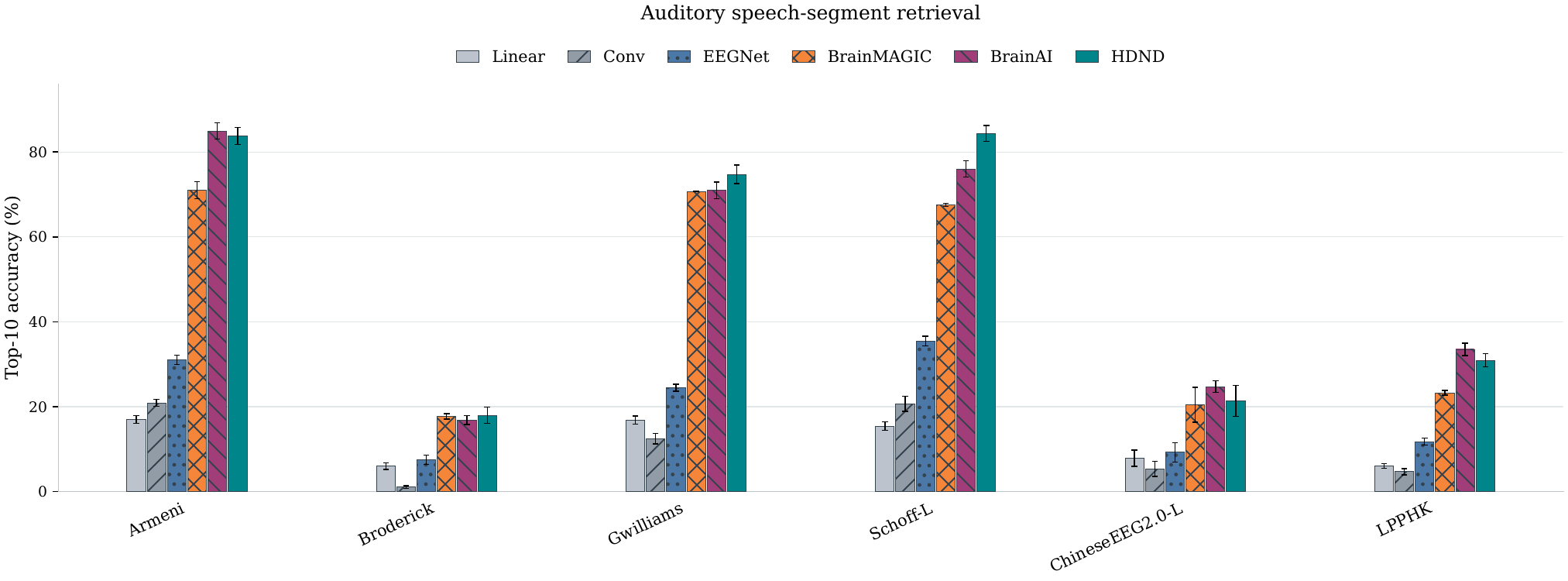}
\caption{Auditory speech-segment retrieval across the six auditory-listening conditions. Bars show participant-averaged Top-10 retrieval accuracy, and error bars show SEM across participants. Reading conditions are excluded because they do not provide a corresponding perceived-speech target. Complete values are reported in Table~\ref{tab:full-speech}.}
\label{fig:speech-condition-bars}
\end{figure*}

Speech-segment retrieval is evaluated only for the six auditory-listening conditions (Fig.~\ref{fig:speech-condition-bars}; complete values are reported in Table~\ref{tab:full-speech}). HDND is higher than BrainAI on Broderick, Gwilliams, and Schoff-L, and lower on Armeni, ChineseEEG2.0-L, and LPPHK. The mean difference is 1.03\pp\ and the median difference is $-0.04\pp$, showing that the speech-segment comparison is substantially less consistent than the word comparison. ChineseEEG2.0-L is especially informative: it favors HDND for word/character and sentence retrieval but favors BrainAI for speech-segment retrieval. This dissociation indicates that the three endpoints probe different properties of the decoded representation.

\begin{table}[H]
\centering
\caption{Descriptive HDND--BrainAI differences across retrieval endpoints. Mean, median, and range summarize equally weighted condition-level differences in percentage points (pp). Speech includes the six auditory-listening conditions only.}
\label{tab:endpoint-comparison}
\small
\begin{tabular}{lrrrr}
\toprule
Endpoint & Positive / total & Mean difference (pp)
& Median difference (pp) & Range (pp) \\
\midrule
Word/character & 9/9 & +4.09 & +3.84 & +0.23 to +9.12 \\
Sentence & 8/9 & +4.19 & +4.55 & $-4.44$ to +13.60 \\
Speech segment & 3/6 & +1.03 & $-0.04$ & $-3.33$ to +8.36 \\
\bottomrule
\end{tabular}
\end{table}

\subsection{Performance across Languages}

\begin{figure*}[t]
\centering
\includegraphics[width=\textwidth]{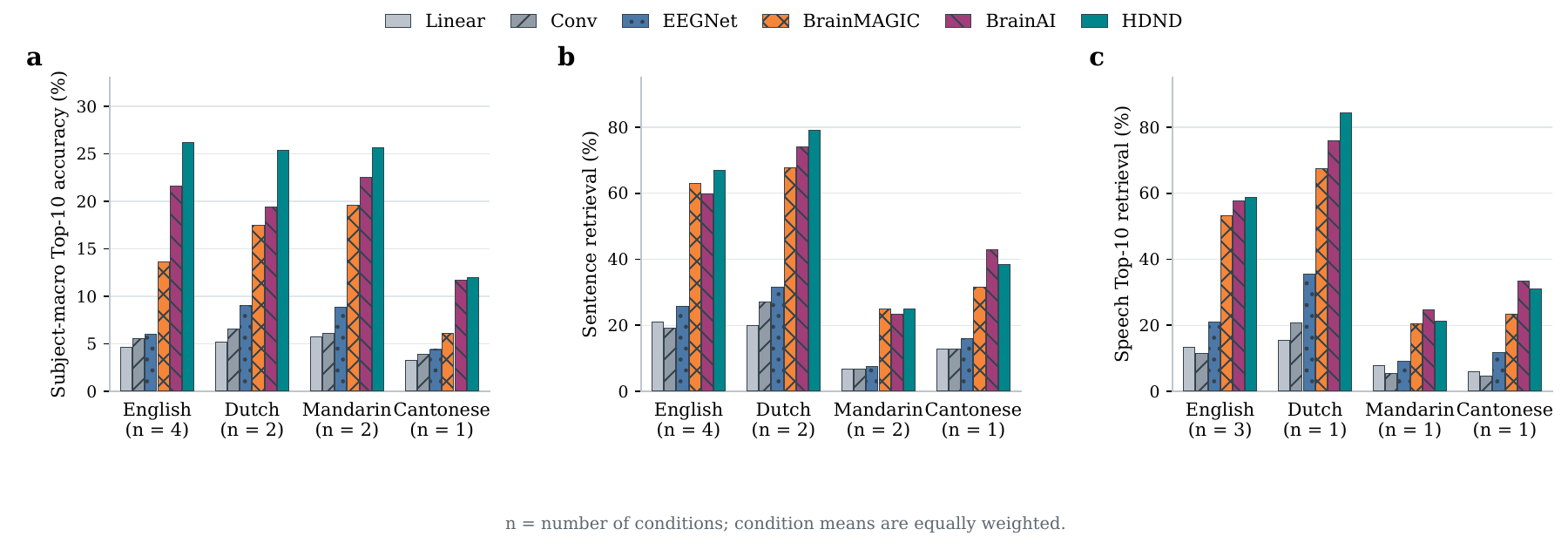}
\caption{Language-level descriptive averages across English, Dutch, Mandarin, and Cantonese conditions. Bars show equally weighted condition means for word/character retrieval, sentence retrieval, and auditory speech-segment retrieval.}
\label{fig:language-averages}
\end{figure*}

Descriptive grouping by language preserves a positive HDND--BrainAI word/character-retrieval difference for the English, Dutch, Mandarin, and Cantonese conditions represented in the primary benchmark. The corresponding descriptive language-level averages are reported in Table~\ref{tab:language-summary}. The average lexical-retrieval differences are larger for the included English and Dutch conditions than for the Mandarin and Cantonese conditions. Sentence-level grouping is less uniform because the Cantonese result is determined by LPPHK alone, while speech-segment comparisons include listening conditions only. Language is also confounded with device, task, stimuli, cohort, and text encoder. These summaries therefore establish coverage across language-diverse datasets, not a language or writing-system effect. Because a separate model is fitted for each condition, they likewise do not demonstrate cross-language transfer. The language-level descriptive averages are visualized in Fig.~\ref{fig:language-averages}.

\subsection{Multilingual Examples}

\begin{figure*}[t]
\centering
\includegraphics[width=\textwidth]{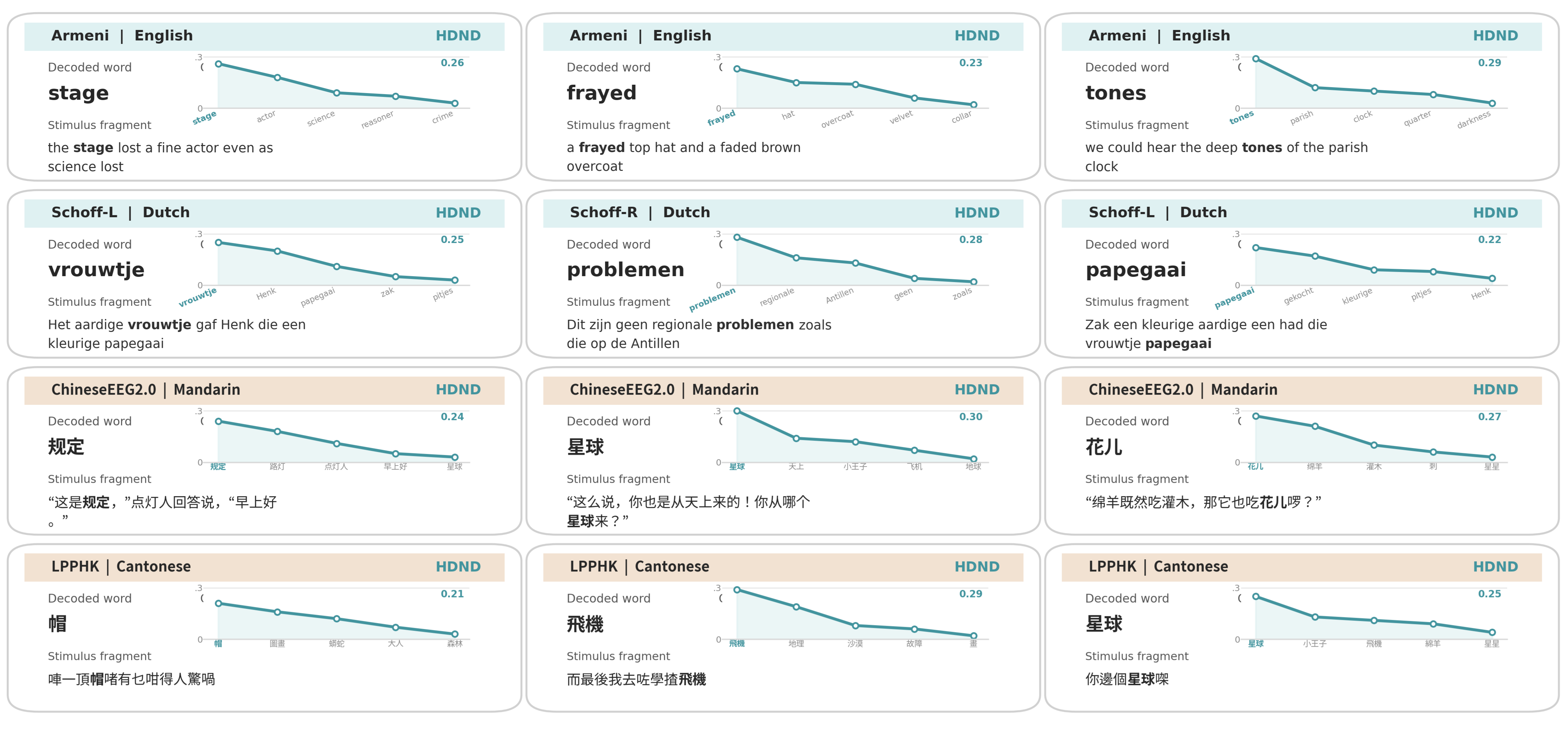}
\caption{Representative held-out test-set word/character-retrieval examples for English, Dutch, Mandarin, and Cantonese. Bold text marks the target lexical unit in each stimulus fragment, and curves show candidate-score orderings produced directly by HDND for the displayed trials. Each panel corresponds to a recorded held-out EEG/MEG trial. The figure is provided for qualitative illustration only and does not define an additional quantitative evaluation.}
\label{fig:examples}
\end{figure*}

Figure~\ref{fig:examples} illustrates the retrieval interface for English, Dutch, Mandarin, and Cantonese. Each panel pairs a selected lexical unit with its stimulus fragment and a short candidate ordering in the original script. The examples illustrate the output format; their displayed scores are not used to compute accuracy or to support language-specific conclusions.

\section{Discussion}
\label{sec:discussion}

\subsection{Hierarchical Refinement of Non-Invasive Word Decoding}
This study asks whether a contextual neural word prediction can be improved by restoring information that may be attenuated during successive temporal encoding and sentence-level aggregation. HDND addresses this problem with a deliberately constrained refinement: intermediate features from two encoder depths are summarized at three temporal resolutions, selected by a contextual semantic query, and returned to the semantic prediction through a bounded residual. The resulting architecture retains the advantages of a geometry-aware, multi-participant neural encoder while exposing local and intermediate-timescale evidence to the final retrieval decision.

The primary result is a consistent descriptive advantage for word retrieval. HDND is higher than BrainAI in all nine matched conditions, spanning EEG and MEG, listening and reading, and four languages. The magnitude of the difference nevertheless varies markedly, from 0.23\pp\ on LPPHK to 9.12\pp\ on Armeni. This variability is scientifically informative: it argues against treating hierarchical refinement as a uniform multiplicative improvement and instead suggests that its value depends on the information retained by the contextual representation, the signal quality of the recording, and the structure of the evaluation protocol.

\subsection{Intermediate Evidence and Contextual Calibration}

The component analyses clarify how the same architecture can help for different reasons. Removing the temporal pathway through the base-only control tests whether contextual remapping is sufficient, while within-participant shuffling tests whether the pathway depends on the correct neural sample. ChineseEEG2.0-L satisfies both criteria: full HDND exceeds both controls, and disrupting sample identity produces the larger loss. This pattern supports the interpretation that intermediate activity contains retrieval-relevant information not fully expressed in the final contextual vector.

Gwilliams and LPPHK exhibit a different regime. Their shuffled controls remain close to the full model, implying that much of the gain can be obtained by recalibrating the contextual representation. Such recalibration is not merely a nuisance explanation: contextual encoders are optimized under a global alignment objective, whereas retrieval depends on local geometry among candidates. A compact residual module can therefore improve ranking by correcting systematic distortions in that geometry. The dynamic fusion can be understood in the same terms. Early and deeper convolutional stages differ in receptive field and abstraction, but their usefulness need not be constant across trials. Conditioning their weights on the contextual prediction allows the refinement to emphasize the representation most compatible with the current linguistic hypothesis. Because the pooling bins within each depth are concatenated, the model is dynamic across encoder depths rather than across individual temporal resolutions. 

\subsection{Surface Form and Language-Diverse Evaluation}
For reading conditions, character-form supervision introduces a constraint complementary to semantic similarity. Character n-grams share statistical structure across lexical items without requiring a vocabulary-sized output classifier \cite{bojanowski2017subword,zhu2019subword}. In visually presented language, this constraint is well matched to the stimulus because orthographic form is explicit and time locked to word presentation. Mapping the predicted form vector back into the common semantic space also preserves a single retrieval rule at inference. However, alphabetic strings, Dutch compounds, Mandarin segmentation, and Cantonese tokenization impose different distributions over surface units. Moreover, each condition is trained independently, and language covaries with task, device, stimuli, and participant cohort. The multilingual value of the present benchmark is therefore breadth of evaluation: the same architectural principle can be instantiated across diverse language materials. 

\subsection{Possible Path towards Practical Neural Language Decoding}
HDND remains an offline retrieval system rather than a free-form brain-to-text interface. It assumes known word onsets and sentence boundaries, and its bidirectional sentence Transformer uses future as well as past context. Participants also contribute to the training and test partitions, so the experiments evaluate held-out linguistic content rather than zero-shot transfer to unseen participants. Finally, condition-specific training prevents conclusions about transfer to new datasets, devices, or languages. These limitations define concrete next steps. Causal contextual models and automatic segmentation are needed for streaming use; participant-independent adaptation is needed for deployment beyond the training cohort; and controlled, cross-device recordings are needed to separate neural, linguistic, and protocol effects. Within the present setting, however, the results establish that contextual word predictions can be improved by a compact hierarchy that combines intermediate neural evidence, surface-form constraints for reading, and semantic calibration.

\section{Conclusion}
\label{sec:conclusion}
The proposed HDND model refines contextual EEG and MEG-based neural word/character predictions through a multi-resolution temporal pyramid, semantic-query fusion across encoder depths, a bounded residual, and reading conditions can additionally use character-form supervision. Under the nine-condition protocol matched to BrainAI, HDND yields a higher balanced Top-10 word-retrieval point estimate in every condition. Core controls show that full HDND exceeds both the shuffled-pyramid and base-output-only controls across all nine conditions, consistent with contributions from correctly paired intermediate features and contextual remapping. Full-vocabulary, sentence, and auditory speech-segment analyses further show that the magnitude and direction of the HDND--BrainAI difference vary across retrieval endpoints and candidate-space definitions. These findings position hierarchical residual refinement as a useful framework for investigating how intermediate neural evidence, surface form, and semantics jointly support word retrieval from heterogeneous non-invasive recordings.

\section*{Acknowledgments}

The authors thank the creators and participants of the public datasets used in this study.

\section*{Funding}

This work was supported by The Hong Kong Polytechnic University Departmental Collaborative Research Fund (Project ID: P0056428), an internal grant from The Hong Kong Polytechnic University (Project ID: P0048377), The Hong Kong Polytechnic University Collaborative Research with World-leading Research Groups Fund (Project ID: P0058097) and Research Grants Council Collaborative Research Fund (Project ID: C5033-24G).

\clearpage
\appendix
\raggedbottom

% ============================================================
% Appendix numbering
% ============================================================
\setcounter{table}{0}
\setcounter{figure}{0}
\renewcommand{\thetable}{A\arabic{table}}
\renewcommand{\thefigure}{A\arabic{figure}}

\section{Appendix}
\label{app:full-results}

The supplementary analyses use the same primary condition definitions as the
main benchmark. The word/character- and sentence-retrieval analyses contain
nine conditions: Armeni, Broderick, Gwilliams, Schoff-L, Schoff-R,
ChineseEEG2.0-L, ChineseEEG2.0-R, LPPHK, and Nieuwland. Auditory
speech-segment retrieval is restricted to the six conditions in which
participants listened to speech: Armeni, Broderick, Gwilliams, Schoff-L,
ChineseEEG2.0-L, and LPPHK.

Unless explicitly stated otherwise, supplementary analyses retain these
condition names and ordering.

% ============================================================
\subsection{Primary Word/Character-Retrieval Results}
% ============================================================

\begin{table}[H]
\centering
\caption{Primary matched benchmark: participant-averaged balanced Top-10 word/character retrieval over 250 candidates. Entries are mean $\pm$ standard error of the mean (SEM) (\%). Arm., Brod., Gwil., Sch.-L/R, Chn.-L/R, and Nieu. denote Armeni, Broderick, Gwilliams, Schoff-L/R, ChineseEEG2.0-L/R, and Nieuwland, respectively.}
\label{tab:full-word250}
\scriptsize
\resizebox{\textwidth}{!}{%
\begin{tabular}{lccccccccc}
\toprule
Method & Arm. & Brod. & Gwil. & Sch.-L & Sch.-R
& Chn.-L & Chn.-R & LPPHK & Nieu. \\
\midrule
Linear
& $4.60 \pm 0.37$
& $4.10 \pm 0.70$
& $4.91 \pm 0.43$
& $5.20 \pm 0.62$
& $5.10 \pm 0.50$
& $5.49 \pm 0.13$
& $5.99 \pm 0.27$
& $3.30 \pm 0.69$
& $4.81 \pm 2.01$ \\

Conv
& $5.71 \pm 0.51$
& $5.20 \pm 0.54$
& $5.98 \pm 0.57$
& $6.30 \pm 0.45$
& $6.80 \pm 0.64$
& $6.21 \pm 1.21$
& $5.91 \pm 2.21$
& $3.90 \pm 0.52$
& $5.24 \pm 0.07$ \\

EEGNet
& $8.10 \pm 0.58$
& $4.70 \pm 0.57$
& $5.30 \pm 0.30$
& $7.30 \pm 0.51$
& $10.78 \pm 0.44$
& $8.21 \pm 0.35$
& $9.51 \pm 1.35$
& $4.40 \pm 0.57$
& $5.91 \pm 2.79$ \\

BrainMAGIC
& $16.84 \pm 0.53$
& $7.10 \pm 0.73$
& $9.30 \pm 0.48$
& $9.40 \pm 0.36$
& $25.57 \pm 0.78$
& $21.53 \pm 0.96$
& $17.67 \pm 2.29$
& $6.10 \pm 0.71$
& $21.31 \pm 1.04$ \\

BrainAI
& $38.07 \pm 0.97$
& $16.18 \pm 0.23$
& $16.81 \pm 0.21$
& $14.22 \pm 0.55$
& $24.67 \pm 0.59$
& $25.75 \pm 0.29$
& $19.21 \pm 0.51$
& $11.72 \pm 0.16$
& $15.31 \pm 1.05$ \\

HDND
& $47.19 \pm 0.93$
& $19.31 \pm 1.03$
& $18.01 \pm 0.19$
& $20.58 \pm 0.77$
& $30.19 \pm 0.79$
& $29.59 \pm 0.34$
& $21.74 \pm 0.28$
& $11.95 \pm 0.14$
& $20.18 \pm 2.75$ \\
\bottomrule
\end{tabular}%
}
\end{table}

The condition ordering in Table~\ref{tab:full-word250} is identical to that used in the primary word/character-retrieval figure. HDND has a higher balanced Top-10 point estimate than BrainAI in all nine conditions.

% ============================================================
\subsection{Primary Sentence-Retrieval Results}
% ============================================================

\begin{table}[H]
\centering
\caption{Primary matched benchmark: participant-averaged sentence Top-10 retrieval. Entries are mean $\pm$ SEM (\%). Abbreviations and condition order match Table~\ref{tab:full-word250}.}
\label{tab:full-sentence}
\scriptsize
\resizebox{\textwidth}{!}{%
\begin{tabular}{lccccccccc}
\toprule
Method & Arm. & Brod. & Gwil. & Sch.-L & Sch.-R & Chn.-L & Chn.-R & LPPHK & Nieu. \\
\midrule
Linear
& $27.87 \pm 2.05$
& $12.67 \pm 1.04$
& $25.62 \pm 0.91$
& $20.99 \pm 1.14$
& $19.08 \pm 1.13$
& $6.23 \pm 0.15$
& $7.03 \pm 1.02$
& $12.91 \pm 1.33$
& $18.25 \pm 1.15$ \\

Conv
& $28.94 \pm 1.04$
& $10.18 \pm 1.18$
& $18.77 \pm 0.56$
& $28.64 \pm 1.85$
& $25.58 \pm 4.23$
& $7.47 \pm 0.19$
& $6.07 \pm 1.19$
& $12.77 \pm 1.23$
& $18.33 \pm 1.42$ \\

EEGNet
& $36.57 \pm 1.38$
& $10.17 \pm 1.12$
& $33.29 \pm 1.75$
& $30.02 \pm 1.16$
& $32.89 \pm 2.83$
& $7.01 \pm 0.56$
& $8.01 \pm 1.44$
& $16.08 \pm 1.14$
& $22.97 \pm 1.21$ \\

BrainMAGIC
& $79.02 \pm 2.12$
& $26.92 \pm 1.22$
& $77.13 \pm 2.31$
& $70.46 \pm 1.32$
& $65.12 \pm 1.81$
& $27.85 \pm 3.24$
& $22.15 \pm 2.97$
& $31.66 \pm 1.53$
& $68.41 \pm 1.75$ \\

BrainAI
& $80.61 \pm 2.04$
& $23.56 \pm 1.28$
& $79.82 \pm 1.53$
& $76.51 \pm 0.35$
& $71.47 \pm 0.35$
& $23.65 \pm 1.49$
& $22.89 \pm 0.79$
& $42.96 \pm 1.27$
& $55.01 \pm 1.78$ \\

HDND
& $86.83 \pm 2.17$
& $37.16 \pm 2.11$
& $86.09 \pm 2.15$
& $81.97 \pm 2.01$
& $76.02 \pm 2.39$
& $25.66 \pm 2.04$
& $24.15 \pm 1.96$
& $38.52 \pm 1.83$
& $57.79 \pm 1.84$ \\
\bottomrule
\end{tabular}%
}
\end{table}

HDND has a higher sentence-retrieval point estimate than BrainAI in eight of the nine conditions. The corresponding HDND--BrainAI differences are $+6.22\pp$, $+13.60\pp$, $+6.27\pp$, $+5.46\pp$, $+4.55\pp$, $+2.01\pp$, $+1.26\pp$, $-4.44\pp$, and $+2.78\pp$. LPPHK is the only primary condition in which the BrainAI point estimate exceeds HDND.

% ============================================================
\subsection{Auditory Speech-Segment Retrieval}
% ============================================================

\begin{table}[H]
\centering
\caption{Matched auditory speech-segment Top-10 retrieval for the six listening conditions. Entries are participant means $\pm$ SEM (\%). Reading and reading-aloud conditions are excluded because they do not contain a simultaneously perceived speech target.}
\label{tab:full-speech}
\scriptsize
\resizebox{\textwidth}{!}{%
\begin{tabular}{lcccccc}
\toprule
Method & Armeni & Broderick & Gwilliams & Schoff-L & ChineseEEG2.0-L & LPPHK \\
\midrule
Linear
& $16.97 \pm 0.87$
& $5.94 \pm 0.78$
& $16.84 \pm 0.92$
& $15.37 \pm 0.99$
& $7.86 \pm 1.91$
& $6.02 \pm 0.58$ \\

Conv
& $20.88 \pm 0.79$
& $1.04 \pm 0.31$
& $12.42 \pm 1.23$
& $20.63 \pm 1.77$
& $5.31 \pm 1.74$
& $4.66 \pm 0.73$ \\

EEGNet
& $31.03 \pm 1.06$
& $7.49 \pm 1.12$
& $24.47 \pm 0.83$
& $35.46 \pm 1.13$
& $9.22 \pm 2.26$
& $11.75 \pm 0.82$ \\

BrainMAGIC
& $70.99 \pm 1.99$
& $17.71 \pm 0.63$
& $70.71 \pm 0.07$
& $67.53 \pm 0.39$
& $20.43 \pm 4.12$
& $23.25 \pm 0.56$ \\

BrainAI
& $84.96 \pm 1.94$
& $16.84 \pm 1.03$
& $70.97 \pm 1.94$
& $76.02 \pm 1.91$
& $24.71 \pm 1.35$
& $33.49 \pm 1.43$ \\

HDND
& $83.76 \pm 2.03$
& $17.96 \pm 1.88$
& $74.73 \pm 2.18$
& $84.38 \pm 1.88$
& $21.38 \pm 3.67$
& $30.93 \pm 1.54$ \\
\bottomrule
\end{tabular}%
}
\end{table}

HDND is higher than BrainAI on Broderick, Gwilliams, and Schoff-L and lower on Armeni, ChineseEEG2.0-L, and LPPHK. The six condition-level differences are $-1.20\pp$, $+1.12\pp$, $+3.76\pp$, $+8.36\pp$, $-3.33\pp$, and $-2.56\pp$. Their mean is $+1.03\pp$ and their median is $-0.04\pp$.

% ============================================================
\subsection{Language-Level Descriptive Summary}
% ============================================================

For descriptive language-level summaries, condition means are equally
weighted rather than pooled over participants. English contains four primary
conditions for word/character and sentence retrieval (Armeni, Broderick,
Gwilliams, and Nieuwland), Dutch contains two (Schoff-L and Schoff-R),
Mandarin contains two (ChineseEEG2.0-L and ChineseEEG2.0-R), and Cantonese
is represented by LPPHK alone. For auditory speech-segment retrieval, only
listening conditions are included: three English conditions and one condition
each for Dutch, Mandarin, and Cantonese.

\begin{table}[H]
\centering
\caption{Descriptive language-level averages derived from the primary
condition means. Conditions are equally weighted within language. Values
are percentages. Speech-segment results include only auditory-listening
conditions.}
\label{tab:language-summary}

\footnotesize
\setlength{\tabcolsep}{3.5pt}

\begin{tabular}{llrrrrrr}
\toprule
Endpoint & Language & Linear & Conv & EEGNet & BrainMAGIC & BrainAI & HDND \\
\midrule

\multirow{4}{*}{Word/character}
& English   & 4.60 & 5.53 & 6.00 & 13.64 & 21.59 & 26.17 \\
& Dutch     & 5.15 & 6.55 & 9.04 & 17.48 & 19.44 & 25.38 \\
& Mandarin  & 5.74 & 6.06 & 8.86 & 19.60 & 22.48 & 25.66 \\
& Cantonese & 3.30 & 3.90 & 4.40 & 6.10 & 11.72 & 11.95 \\

\midrule

\multirow{4}{*}{Sentence}
& English   & 21.10 & 19.06 & 25.75 & 62.87 & 59.75 & 66.97 \\
& Dutch     & 20.03 & 27.11 & 31.46 & 67.79 & 73.99 & 79.00 \\
& Mandarin  & 6.63 & 6.77 & 7.51 & 25.00 & 23.27 & 24.90 \\
& Cantonese & 12.91 & 12.77 & 16.08 & 31.66 & 42.96 & 38.52 \\

\midrule

\multirow{4}{*}{Speech segment}
& English   & 13.25 & 11.45 & 21.00 & 53.14 & 57.59 & 58.82 \\
& Dutch     & 15.37 & 20.63 & 35.46 & 67.53 & 76.02 & 84.38 \\
& Mandarin  & 7.86 & 5.31 & 9.22 & 20.43 & 24.71 & 21.38 \\
& Cantonese & 6.02 & 4.66 & 11.75 & 23.25 & 33.49 & 30.93 \\

\bottomrule
\end{tabular}
\end{table}

These language-level values are descriptive summaries of the condition-level
results and should not be interpreted as estimates of a language effect.
Language is confounded with dataset, recording device, task, stimuli, and
participant cohort, and Cantonese is represented by a single condition.

% ============================================================
\subsection{Full-Vocabulary Retrieval}
\label{app:full-vocabulary-analysis}
% ============================================================

\begin{table}[H]
\centering
\caption{Full-vocabulary Top-10 word/character retrieval across the nine primary conditions. Values are token-micro / pooled word-macro percentages from the validation-selected model for each condition using the same trained backbone as the matched BrainAI comparison.}
\label{tab:full-vocabulary}
\small
\begin{tabular}{lcc}
\toprule
Condition & BrainAI micro / macro & HDND micro / macro \\
\midrule
Armeni
& 15.80 / 5.80
& 22.40 / 6.85 \\
Broderick
& 2.76 / 1.66
& 4.84 / 2.13 \\
Gwilliams
& 13.12 / 4.61
& 16.76 / 5.04 \\
Schoff-L
& 7.90 / 3.10
& 12.60 / 3.75 \\
Schoff-R
& 12.60 / 4.40
& 17.80 / 5.00 \\
ChineseEEG2.0-L
& 10.98 / 2.46
& 17.11 / 3.03 \\
ChineseEEG2.0-R
& 11.34 / 1.83
& 16.17 / 2.21 \\
LPPHK
& 7.03 / 4.91
& 8.93 / 5.30 \\
Nieuwland
& 6.18 / 2.72
& 9.97 / 3.67 \\
\bottomrule
\end{tabular}
\end{table}

% ============================================================
\subsection{Core Component Analysis}
\label{app:core-components}
% ============================================================

The core component analysis follows the same nine condition definitions as the primary word/character-retrieval benchmark. The shuffled-pyramid control retains the hierarchical pathway while disrupting the correspondence between each example and its intermediate pyramid representation. The base-output-only control removes the temporal-pyramid pathway and learns the residual correction from the contextual representation alone.

\begin{table}[H]
\centering
\caption{Core component analysis across the nine primary word/character-retrieval conditions. Values are participant-averaged balanced Top-10 accuracy (\%).}
\label{tab:core-ablations}
\small
\begin{tabular}{lrrr}
\toprule
Condition & HDND & Shuffled-pyramid & Base-output-only \\
\midrule
Armeni
& 47.19 & 43.89 & 44.36 \\
Broderick
& 19.31 & 17.70 & 18.31 \\
Gwilliams
& 18.01 & 16.52 & 17.07 \\
Schoff-L
& 20.58 & 18.52 & 19.76 \\
Schoff-R
& 30.19 & 27.77 & 28.53 \\
ChineseEEG2.0-L
& 29.59 & 27.14 & 28.05 \\
ChineseEEG2.0-R
& 21.74 & 19.93 & 20.61 \\
LPPHK
& 11.95 & 10.96 & 11.33 \\
Nieuwland
& 20.18 & 18.51 & 19.13 \\
\bottomrule
\end{tabular}
\end{table}

% ============================================================
\subsection{Implementation Details}
\label{app:implementation}
% ============================================================

\begin{table}[H]
\centering
\caption{Configuration and validation search space of the matched word/character-retrieval experiments.}
\label{tab:hyperparameters}
\small
\begin{tabular}{ll}
\toprule
Component & Setting \\
\midrule
Input window & 0--3~s from word onset \\
Initial filtering and resampling & 0.1--40~Hz band-pass; resampled to 50~Hz with anti-alias filtering \\
Scaling & Robust scaling; clamp $[-5,5]$ \\
Sensor-position encoding & Fourier coordinates; 2,048 dimensions \\
Merged latent channels & 270 \\
Initial projection & 512 channels \\
Temporal encoder & Five convolutional stages; 160 channels in stages 1--4 and 1,024 in stage 5 \\
Pyramid inputs & GLU outputs after convolutional stages 2 and 4 \\
Pooling resolutions & 1, 2, and 4 bins \\
Pyramid input width & 1,120 features per depth \\
Projected width & 64 or 128 \\
Semantic representation & 1,024 dimensions \\
Form representation & 300 dimensions \\
Character n-grams & Unique boundary-marked lengths 2--6; fixed seeded 300-D Gaussian projection \\
Sentence Transformer & 16 layers; 16 heads \\
Pre-centering residual bound & 0.15 or 0.25 \\
Residual strength & 0.125, 0.25, 0.5, or 1.0 \\
Optimizer & AdamW; weight decay 0.01 \\
Batch / schedule & 1,024; cosine decay; gradient norm 1.0 \\
\bottomrule
\end{tabular}
\end{table}

\paragraph{Training configurations.} The 64-wide search family uses a learning rate of $3\times10^{-4}$ and a KL weight of 0.5. The 128-wide family uses a learning rate of $2\times10^{-4}$, a KL weight of 0.25, a form-cosine weight of 0.1 when form supervision is enabled, and a within-participant pairing weight of 0.3. Unless otherwise specified, the frequency-weight exponent is $-0.25$. Refined Broderick configurations use a learning rate of $10^{-4}$, a KL weight of 0.5, and a residual bound of 0.15. Warm-start configurations continue a previously trained HDND refinement module at $10^{-5}$ or $3\times10^{-5}$. Form supervision is considered only for Schoff-R, ChineseEEG2.0-R, and Nieuwland reading conditions and is disabled for every listening condition.

\paragraph{Validation selection.} Checkpoint and configuration selection is based exclusively on validation-set performance using token-micro and word-macro accuracy. Targeted searches additionally consider full-vocabulary validation performance. The validation-selected HDND refinement model is then applied unchanged to the test set. Test results and BrainAI test predictions do not enter model selection, and every reported HDND value is produced by a trained refinement model applied to the same frozen contextual backbone used for the matched BrainAI comparison.

\paragraph{Search procedure.} Each reported condition uses one trained contextual backbone shared by the matched BrainAI and HDND evaluations. HDND refinement configurations and checkpoints are selected using validation data only, and random initialization is not treated as an additional inferential unit in the reported test uncertainty. Reported SEM values quantify variability across participants, consistent with the primary evaluation protocol. The semantic-only 64-wide refinement module contains 274,432 trainable parameters, and adding the form pathway gives 600,832 parameters. The 128-wide module with the form pathway contains 894,464 trainable parameters. These counts exclude the neural encoder trained during the first stage.
\bibliographystyle{unsrtnat}
\bibliography{references}

\end{document}